\pdfoutput=1

\documentclass[11pt]{article}

\usepackage{xcolor}

\newcommand{\customannotate}[3]{}

\usepackage{acl}

\usepackage{times}
\usepackage{latexsym}
\usepackage{hyperref}
\usepackage{float}
\usepackage{multicol}

\usepackage[T1]{fontenc}

\usepackage[utf8]{inputenc}

\usepackage{microtype}

\usepackage{inconsolata}

\usepackage{graphicx}
\title{Question-Analysis Prompting Improves LLM Performance in Reasoning Tasks}

\author{Dharunish Yugeswardeenoo \hspace{1cm} Kevin Zhu \hspace{1cm} Sean O'Brien \\
        Algoverse AI Research \\
        \texttt{dharyugi@gmail.com, kevin@algoverseacademy.com}}

\begin{document}
\maketitle
\begin{sloppypar}
\begin{abstract}

Although LLMs have the potential to transform many fields, they still underperform humans in reasoning tasks. Existing methods induce the model to produce step-by-step calculations, but this research explores the question: Does making the LLM analyze the question improve its performance? We propose a novel prompting strategy called Question Analysis Prompting (QAP), in which the model is prompted to explain the question in \textit{n} words before solving. The value of \textit{n} influences the length of response generated by the model. QAP is evaluated on GPT-3.5 Turbo and GPT-4 Turbo on arithmetic datasets GSM8K, AQuA, and SAT and commonsense dataset StrategyQA. QAP is compared with other state-of-the-art prompts including chain-of-thought (CoT), Plan and Solve Prompting (PS+) and Take A Deep Breath (TADB). QAP outperforms all state-of-the-art prompts on AQuA and SAT datasets on both GPT-3.5 and GPT-4. QAP consistently ranks among the top-2 prompts on 75\% of the tests. A key factor of QAP performance can be attributed to response length, where detailed responses are beneficial when answering harder questions, but can negatively affect easy questions.

\end{abstract}
\section{Introduction}
 
    Large language models (LLMs) have recently shown rapid improvement across a host of standard natural language processing (NLP) tasks, including arithmetic, commonsense and symbolic reasoning. \citep{brown2020language}
    Although these models show improved ability to understand and generate text \citep{OpenAI2023GPT4TR}, their performance can still be further improved.
    One solution is to encourage the model to think step-by-step. Using chain-of-thought prompting \citep{wei2022chain}, LLMs are given Q\&A exemplars which are designed to elicit a structured step-by-step response from the model. Many newly developed strategies meant to improve LLM performance have been focused on sophisticating the model's step-by-step calculation \citep{gu2023systematic}.
    Despite SoTA prompts' remarkable success across various tasks, their accuracies can still be further improved.
    In this work, we explore ways to improve the model reasoning not only in the answer steps, but also how the model interprets the question itself. By making the model to explicitly interpret the question, we maximize its understanding of the question and minimize missed key information.
    This paper introduces Question-Analysis Prompting (QAP), a simple zero-shot prompting strategy that induces the model to first explain the question before solving.
    We include a configurable parameter within the prompt to examine how different word counts affect the quality of a model's response.
    
\section{Prompt Design}
The key principle behind QAP is that the model should reiterate the problem in its own words before solving. The benefit is that the model will be able to first think about what task it is trying to solve before it pursues the answer.
Another principle is that we should be able to control how much the model explains so that we can adapt the prompt to different model sizes and problem complexities.
The specific prompt used is as follows:

\begin{center}
\fbox{%
  \begin{minipage}{0.45\textwidth}
    \centering
    "Explain this problem to me in at least $n$ words.
    
    Then solve for the answer."
  \end{minipage}%
}
\end{center}
In this work, we experiment with \textit{n} = 25, 50, 100, 150, 200.
The versions of these prompts are named QAP$n$.
Although the model is not constrained to generating fewer than \textit{n} tokens in its summary, we find that the number of tokens in the response correlates strongly with the choice of $n$. The correlation between \textit{n} and median word count is 0.98.
We show specific examples of the impacts of $n$ in~\autoref{fig:nexample}.
\begin{figure*}[ht!]
    \centering
    \includegraphics[width=0.9\textwidth]{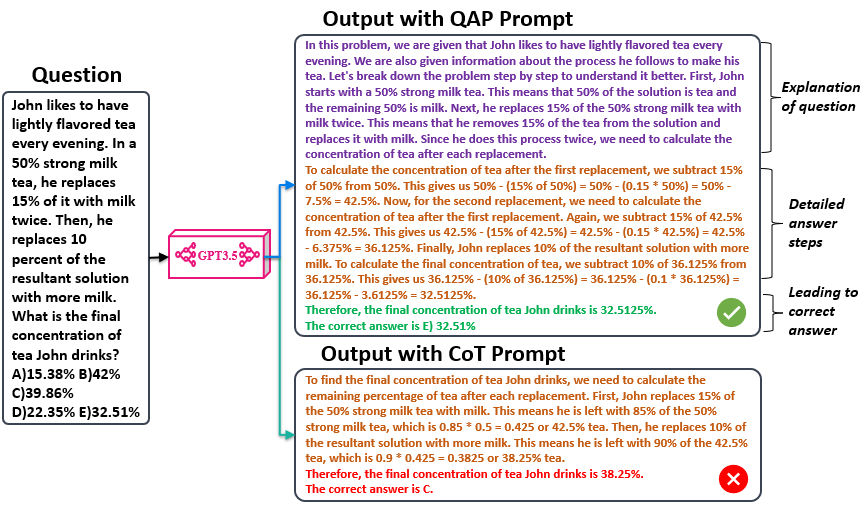}
    \caption{Example of QAP prompting - shows how the prompt triggers explanation of the question followed by an approach to solve the problem, detailed steps, finally leading to correct answer}
    \label{fig:prompt}
\end{figure*}
\section{Prompt Impact}

In~\autoref{fig:prompt}, we highlight the structure of a standard QAP output. First, the model breaks down the question in its own words and provides detailed analysis on each event. Many of the steps highlighted in the explanation were shown in the calculation section. Compared to the CoT output, QAP encourages more sophistication in its response and thus reaches the correct answer.

\section{Experimental Setup}

\subsection{Benchmarks}
We evaluate the effectiveness of QAP on three arithmetic reasoning datasets.
These include grade-school math questions from \textbf{GSM8K} \citep{cobbe2021training}, algebraic word problems from \textbf{AQuA} \citep{ling2017program}, and SAT math problems from \textbf{AGIEval} \citep{zhong2023agieval}.
For commonsense reasoning, we evaluate on open-domain questions that require implicit reasoning, from \textbf{StrategyQA} \citep{geva2021did}. 
We evaluate on the test sets of all benchmarks.

\subsection{Models}
We specifically choose our models to observe the prompts' impacts across differences in model size. The smaller model is {\textbf{GPT-3.5 Turbo }} with version \verb|gpt-3.5-turbo-0613|. Our larger model is \textbf{GPT-4 Turbo} with version \verb|gpt-4-1106-preview| \citep{OpenAI2023GPT4TR}.
For both of the models we used the OpenAI API \footnote{\url{https://platform.openai.com/docs/api-reference/chat}} for running our experiments. The temperature and Top-K sampling was set to 0 to avoid randomness and keep consistency in the model's responses.

\subsection{Prompts}
For all datasets and models, we experiment with different variations of QAP. We utilize \textbf{QAP25}, \textbf{QAP50}, \textbf{QAP100}, \textbf{QAP150}, and \textbf{QAP200}. We compare the performance of QAP with the baseline  (no prompt). Additionally we compare QAP with two different zero-shot prompts: \textbf{TADB} - "Take a deep breath and work on this problem step-by-step" \citep{yang2023large} and \textbf{PS+} (Plan and Solve Plus) \citep{wang2023plan}. Finally we also compare QAP with 8-shot chain-of-thought prompting.

\subsection{Results}

The results for GPT-3.5 Turbo and GPT-4 Turbo are shown in ~\autoref{tab:GPT3.5} and ~\autoref{tab:GPT4} respectively. General word counts are shown in~\autoref{fig:boxplots}.

\textbf{Arithmetic Reasoning:} On GPT-3.5 Turbo, a variant of QAP is the top performer in 2 out of 3 arithmetic tasks. 
QAP shows significant gains on AQuA and SAT.
With GPT-4 Turbo, QAP performs the best in the same 2 out of 3 arithmetic tasks. 
This suggests that QAP may be more beneficial on questions involving algebraic and higher-level problem solving.

\begin{table}
\centering
\begin{tabular}{lclll}
\hline
\textbf{Prompt}& \textbf{GSM8K}&\textbf{AQuA} &\textbf{SAT}&\textbf{StratQA}\\
\hline
Baseline& 78.7& 52.8& 70.9&\textbf{65.1}\\
QAP25& 67.1& 39.4& 35.0&63.1\\
QAP50& 77.8& 50.0& 52.7&61.4\\ 
QAP100& 77.4& 53.9& 75.0&57.1\\ 
QAP150& 78.5& \textbf{59.4}& \textbf{78.6}&53.2\\
QAP200& 76.8& 52.4& 75.0&51.8\\ 
TADB& 78.5& 57.1& 74.5&62.9\\
     CoT& \textbf{79.0}& 53.1& 65.9&59.2\\
 PS+& 74.7& 35.0& 70.9&35.6\\
 \hline
\end{tabular}
\caption{Results for GPT-3.5 Turbo }
\label{tab:GPT3.5}
\end{table}


\begin{table}
\centering
\begin{tabular}{lclll}
\hline
\textbf{Prompt}& \textbf{GSM8K}&\textbf{AQuA} &\textbf{SAT}&\textbf{StratQA}\\
\hline
Baseline& 95.3& 78.7& 96.8&76.3\\
QAP25& 94.8& 77.6& 94.5& 77.6\\
QAP50& 93.4& \textbf{79.1}& 95.9&76.9\\ 
QAP100& 94.6& 75.6& 96.8&77.2\\ 
QAP150& 94.7& 78.0& 97.3&77.6\\
QAP200& 95.0& 76.4& \textbf{98.2}&75.9\\ 
TADB& 95.1& 78.7& 96.8&\textbf{78.0}\\
 CoT& \textbf{95.6}& 74.4& 95.0&75.1\\
 PS+& 94.8& 52.8& 97.3&77.1\\
 \hline
\end{tabular}
\caption{Results for GPT-4 Turbo.}
\label{tab:GPT4}
\end{table}

\textbf{Commonsense Reasoning:}. On StrategyQA, QAP consistently performs second-best when compared to other prompts.
On both models, QAP25 is the highest QAP performer.
This suggests that fewer-word explanations benefit commonsense reasoning. This is because too much explanation can cause the model to confuse a simple answer (shown in~\autoref{fig:comsenseoverex}.
While there is a decline in performance as $n$ increases on the 3.5 model, the larger GPT-4 Turbo model yields similar performances across all QAP variants. 

\section{Analysis}

\textbf{Question Difficulties Based On Baseline Performance:}
Within a given dataset, the difficulty of the individual question may vary.
We propose a method to measure question difficulty based on performance with the baseline prompt.
If the model can answer the problem correctly with the baseline prompt, then we consider the question to be \textit{easy}; otherwise the question is \textit{hard}. We analyze the performance of different prompts across "easy" and "hard" questions. \autoref{tab:arithscore} and \autoref{tab:score} show that QAP consistently outperforms other prompts in the ``hard'' category.

\textbf{Impact Of Word Counts On Question Difficulties:}
QAP generates higher word counts for both ``easy" and ``hard" questions (~\autoref{tab:arith} and ~\autoref{tab:count} ), despite performing lower on ``easy'' questions. Although more step-by-step thought processes are encouraged to avoid mistakes during reasoning, this suggests that \textit{over}-explanation can negatively impact the model (also shown in~~\autoref{fig:OverEx}). Thus, the most suitable word count to solve a problem will vary from task to task; longer explanations are best suited to more complicated questions for which baseline prompting fails.

\textbf{Downsides Of Smaller QAPs:}
Despite high performance on StrategyQA, QAP25 performs poorly on arithmetic datasets (mostly SAT and AQuA) using GPT-3.5 Turbo. Due to a small value of n, the model outputs are unfinished responses (i.e. the model stops midway through its reasoning steps) (shown in~\autoref{fig:unfinresp}). On SAT math, 51\% of responses were incomplete for QAP25.
On AQuA, 19\% of responses were incomplete for QAP25.
\begin{figure*}
    \centering
    \includegraphics[width=0.7\linewidth]{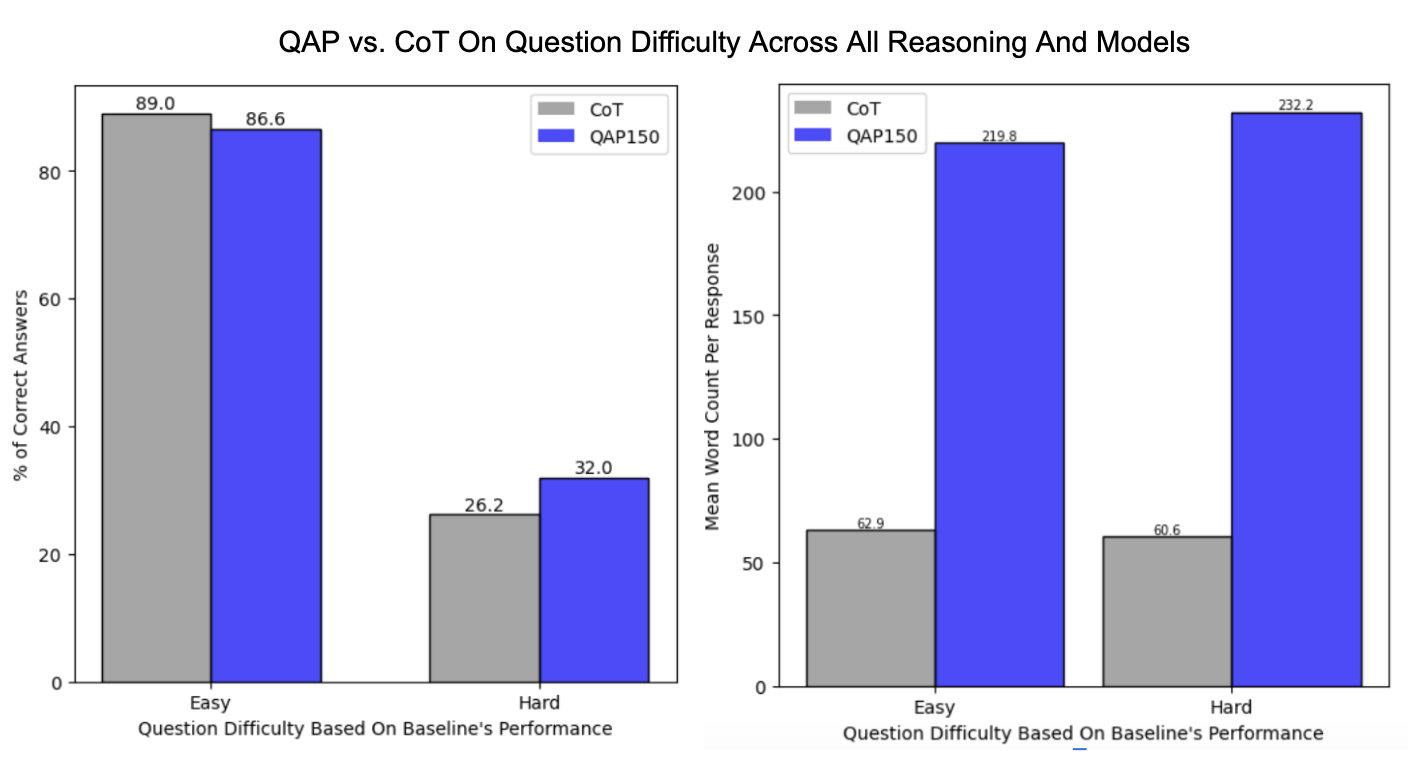}
    \caption{We consider difficulty of the problem based on baseline's results. E.g., an incorrect answer is ``hard'' and a correct answer is ``easy''. Left chart shows accuracy within each difficulty. Right chart shows mean (average) word count for within each difficulty. All results for each prompt are shown in Table \ref{tab:count} and Table \ref{tab:score}}.
    \label{fig:qapvscot}
\end{figure*}
\section{Additional Studies}
\textbf{Placement of the prompt:} In this evaluation, we studied the impact of prompt placement on performance using GSM8K dataset. Two options for prompt placement were considered: Q\_Begin - adding the prompt  before the question, and Q\_End - adding the prompt after the question. Both placements provided similar results on GPT-3.5 and GPT-4. Results shown in the rest of the paper are based on Q\_End.

\textbf{No \textit{N} Constraint}: To test the effectiveness of adding the value of \textit{N}, we first examine the prompt with just the phrase: "Explain this problem to me. Then solve for the answer". However, the model does not explain the question completely and in most cases directly starts solving the question. Its responses are no different than a response which used no given prompt. This shows that explicitly stating the minimum amount of words required is more likely to induce the model to explicitly generate an explanation of the question.

\section{Related Work}

In one-shot and few-shot prompting, the model is given one or more input/output examples which will serve as a demonstration for it to solve the problem using in-context learning \citep{mahabadi2022perfect}. QAP is a zero-shot prompt. In zero-shot prompting the model does not receive exemplars, but is given a specially crafted instruction on how to approach the task \citep{kojima2022large}. 

\textbf{Chain of Thought:}
Chain-of-thought reasoning is a notable few-shot (zero-shot also exists \citep{yang2023large} example in which the model is shown how to express its reasoning steps \citep{wei2022chain}. This approach was highly effective as the model would replicate these exemplars, and their accuracies improved drastically. CoT encouraged the model to think step-by-step, and this concept would be a repeating theme among other zero-shot counterparts. 

\textbf{TADB:}
Among different variants of Zero-Shot CoT, the TADB prompt \citep{yang2023large} was derived using an optimization objective to find instructions that would maximize task accuracy. The eventual prompt was "Take a deep breath, and work on this problem step by step". TADB is an example of how the wording of a prompt can drastically impact responses.  

\textbf{Plan and Solve Prompting Plus:}
Another zero-shot prompt is Plan-and-Solve Prompting \citep{wang2022self}. There were two versions to this prompt. The first simply asked the model devise a plan and solve step-by-step. The second version (PS+) extended the prompt by specifically asking the prompt to extract relevant variables and their corresponding numerals and to calculate intermediate results\textit{.} We used PS+ on our experiments. One difference between PS+ and QAP is that PS+ prompt is more specific to math datasets since it instructs to extract variables, intermediate results, etc., whereas QAP is more general. Also,  PS+ prompts the model to understand the problem, but it is not clear if model should output anything specific to the question itself. In contrast, QAP explicitly instructs the model to explain the problem in \textit{n} words. 

\textbf{Question Decomposition:}
Question Decomposition \citep{Radhakrishnan2023QuestionDI} strategy  causes the model to break down the question by creating sub-questions. The model answers each of these sub-questions and it ties together all the sub-answers into a final answer. It considers two methods for decomposition, Factored Decomposition and CoT Decomposition. In factored decomposition each sub-question is answered in a separate context. CoT decomposition is an intermediate between factored decomposition and CoT. It enforces one context for sub-question, sub-answer and the answer to the original question. The analysis of question decomposition shows reduced bias and ignored reasoning, improves the faithfulness of a model-generated reasoning over CoT while retaining the performance gains of CoT.

\section{Conclusion}
In this paper, we explored the approach of Question-Analysis Prompting to improve LLM accuracy across math and commonsense reasoning. The prompt focuses on how the model interprets the task given, and whether restating the question in its own words can further sophisticate its answer steps.
The ability of this prompting method to perform well in diverse model types, tasks difficulty, and type of tasks seems promising.
We plan to extend this work further by combining QAP with other prompt strategies,  applying decoding strategies and evaluating multi-modal tasks.

\section{Limitations}
There are a few limitations of QAP.
First, LLMs are sensitive to the prompt's word choice, particularly for zero-shot prompts.
As a result so small changes to the prompt wording can impact the model's performance.
For example, the current QAP prompt asks the model to "solve" for the answer.
While this works well for math tasks, it may not be optimal for commonsense tasks.
Secondly, the results in this paper are based on four datasets and a single class of aligned models; further results should evaluate on more diverse and multi-modal datasets, as well as a greater variety of models.
Finally, more robust methods (e.g., based on a classifier) to determine the choice of the parameter $n$ should be investigated to go beyond manual selection.

\section{Ethics}
We experiented on three arithmetic datasets: GSM8K \citep{cobbe2021training},  AQuA \citep{ling2017program}, and AGIEval SAT Math \citep{zhong2023agieval}.
For commonsense reasoning, used StrategyQA \citep{geva2021did}. GSM8K use the MIT License code, while AQUA and StrategyQA use the Apache-2.0 code. 
QAP and the prompts used in this work do not jeopardize the safety of others. They do not include any wording which may deem offensive to any individual or group.

\bibliography{anthology,custom}

\begin{thebibliography}{14}
\expandafter\ifx\csname natexlab\endcsname\relax\def\natexlab#1{#1}\fi

\bibitem[{Brown et~al.(2020)Brown, Mann, Ryder, Subbiah, Kaplan, Dhariwal, Neelakantan, Shyam, Sastry, Askell et~al.}]{brown2020language}
Tom Brown, Benjamin Mann, Nick Ryder, Melanie Subbiah, Jared~D Kaplan, Prafulla Dhariwal, Arvind Neelakantan, Pranav Shyam, Girish Sastry, Amanda Askell, et~al. 2020.
\newblock Language models are few-shot learners.
\newblock \emph{Advances in neural information processing systems}, 33:1877--1901.

\bibitem[{Cobbe et~al.(2021)Cobbe, Kosaraju, Bavarian, Chen, Jun, Kaiser, Plappert, Tworek, Hilton, Nakano et~al.}]{cobbe2021training}
Karl Cobbe, Vineet Kosaraju, Mohammad Bavarian, Mark Chen, Heewoo Jun, Lukasz Kaiser, Matthias Plappert, Jerry Tworek, Jacob Hilton, Reiichiro Nakano, et~al. 2021.
\newblock Training verifiers to solve math word problems.
\newblock \emph{arXiv preprint arXiv:2110.14168}.

\bibitem[{Geva et~al.(2021)Geva, Khashabi, Segal, Khot, Roth, and Berant}]{geva2021did}
Mor Geva, Daniel Khashabi, Elad Segal, Tushar Khot, Dan Roth, and Jonathan Berant. 2021.
\newblock Did aristotle use a laptop? a question answering benchmark with implicit reasoning strategies.
\newblock \emph{Transactions of the Association for Computational Linguistics}, 9:346--361.

\bibitem[{Gu et~al.(2023)Gu, Han, Chen, Beirami, He, Zhang, Liao, Qin, Tresp, and Torr}]{gu2023systematic}
Jindong Gu, Zhen Han, Shuo Chen, Ahmad Beirami, Bailan He, Gengyuan Zhang, Ruotong Liao, Yao Qin, Volker Tresp, and Philip Torr. 2023.
\newblock A systematic survey of prompt engineering on vision-language foundation models.
\newblock \emph{arXiv preprint arXiv:2307.12980}.

\bibitem[{Kojima et~al.(2022)Kojima, Gu, Reid, Matsuo, and Iwasawa}]{kojima2022large}
Takeshi Kojima, Shixiang~Shane Gu, Machel Reid, Yutaka Matsuo, and Yusuke Iwasawa. 2022.
\newblock Large language models are zero-shot reasoners.
\newblock \emph{Advances in neural information processing systems}, 35:22199--22213.

\bibitem[{Ling et~al.(2017)Ling, Yogatama, Dyer, and Blunsom}]{ling2017program}
Wang Ling, Dani Yogatama, Chris Dyer, and Phil Blunsom. 2017.
\newblock Program induction by rationale generation: Learning to solve and explain algebraic word problems.
\newblock \emph{arXiv preprint arXiv:1705.04146}.

\bibitem[{Mahabadi et~al.(2022)Mahabadi, Zettlemoyer, Henderson, Saeidi, Mathias, Stoyanov, and Yazdani}]{mahabadi2022perfect}
Rabeeh~Karimi Mahabadi, Luke Zettlemoyer, James Henderson, Marzieh Saeidi, Lambert Mathias, Veselin Stoyanov, and Majid Yazdani. 2022.
\newblock Perfect: Prompt-free and efficient few-shot learning with language models.
\newblock \emph{arXiv preprint arXiv:2204.01172}.

\bibitem[{OpenAI(2023)}]{OpenAI2023GPT4TR}
OpenAI. 2023.
\newblock \href {https://api.semanticscholar.org/CorpusID:257532815} {Gpt-4 technical report}.
\newblock \emph{ArXiv}, abs/2303.08774.

\bibitem[{Radhakrishnan et~al.(2023)Radhakrishnan, Nguyen, Chen, Chen, Denison, Hernandez, Durmus, Hubinger, Kernion, Lukovsiut.e, Cheng, Joseph, Schiefer, Rausch, McCandlish, Showk, Lanham, Maxwell, Chandrasekaran, Hatfield-Dodds, Kaplan, Brauner, Bowman, and Perez}]{Radhakrishnan2023QuestionDI}
Ansh Radhakrishnan, Karina Nguyen, Anna Chen, Carol Chen, Carson~E. Denison, Danny Hernandez, Esin Durmus, Evan Hubinger, John Kernion, Kamil.e Lukovsiut.e, Newton Cheng, Nicholas Joseph, Nicholas Schiefer, Oliver Rausch, Sam McCandlish, Sheer~El Showk, Tamera Lanham, Tim Maxwell, Venkat Chandrasekaran, Zac Hatfield-Dodds, Jared Kaplan, Janina Brauner, Sam Bowman, and Ethan Perez. 2023.
\newblock \href {https://api.semanticscholar.org/CorpusID:259980634} {Question decomposition improves the faithfulness of model-generated reasoning}.
\newblock \emph{ArXiv}, abs/2307.11768.

\bibitem[{Wang et~al.(2023)Wang, Xu, Lan, Hu, Lan, Lee, and Lim}]{wang2023plan}
Lei Wang, Wanyu Xu, Yihuai Lan, Zhiqiang Hu, Yunshi Lan, Roy Ka-Wei Lee, and Ee-Peng Lim. 2023.
\newblock Plan-and-solve prompting: Improving zero-shot chain-of-thought reasoning by large language models.
\newblock \emph{arXiv preprint arXiv:2305.04091}.

\bibitem[{Wang et~al.(2022)Wang, Wei, Schuurmans, Le, Chi, Narang, Chowdhery, and Zhou}]{wang2022self}
Xuezhi Wang, Jason Wei, Dale Schuurmans, Quoc Le, Ed~Chi, Sharan Narang, Aakanksha Chowdhery, and Denny Zhou. 2022.
\newblock Self-consistency improves chain of thought reasoning in language models.
\newblock \emph{arXiv preprint arXiv:2203.11171}.

\bibitem[{Wei et~al.(2022)Wei, Wang, Schuurmans, Bosma, Xia, Chi, Le, Zhou et~al.}]{wei2022chain}
Jason Wei, Xuezhi Wang, Dale Schuurmans, Maarten Bosma, Fei Xia, Ed~Chi, Quoc~V Le, Denny Zhou, et~al. 2022.
\newblock Chain-of-thought prompting elicits reasoning in large language models.
\newblock \emph{Advances in Neural Information Processing Systems}, 35:24824--24837.

\bibitem[{Yang et~al.(2023)Yang, Wang, Lu, Liu, Le, Zhou, and Chen}]{yang2023large}
Chengrun Yang, Xuezhi Wang, Yifeng Lu, Hanxiao Liu, Quoc~V Le, Denny Zhou, and Xinyun Chen. 2023.
\newblock Large language models as optimizers.
\newblock \emph{arXiv preprint arXiv:2309.03409}.

\bibitem[{Zhong et~al.(2023)Zhong, Cui, Guo, Liang, Lu, Wang, Saied, Chen, and Duan}]{zhong2023agieval}
Wanjun Zhong, Ruixiang Cui, Yiduo Guo, Yaobo Liang, Shuai Lu, Yanlin Wang, Amin Saied, Weizhu Chen, and Nan Duan. 2023.
\newblock Agieval: A human-centric benchmark for evaluating foundation models.
\newblock \emph{arXiv preprint arXiv:2304.06364}.

\end{thebibliography}

\appendix

\section{Appendix}

\subsection{Analysis of Accuracy Based On Question Difficulty }

Performance of prompts on problems categorized into easy and hard - where easy problems are those where baseline prompt leads to a correct answer and hard problems are those where baseline prompt leads to a wrong answer. For each category the \% of correct answers are calculated by number of correct answers(per prompt) over the total number of problems in that category (easy or hard)

\begin{table}[ht!]
\centering
\begin{tabular}{lcl}
\hline
\textbf{Prompt}& \textbf{Easy}&\textbf{Hard}\\
QAP25& 84.7& 30.1\\
QAP50& 90.0& 36.7\\ 
QAP100& 91.5& 39.5\\ 
QAP150& 92.3& 43.2\\
QAP200& 91.1& 41.3\\ 
TADB& 93.6& 34.9\\
 CoT& 92.6& 35.0\\
 PS+& 88.2& 31.5\\
\end{tabular}
\caption{Accuracy for Arithmetic Reasoning}
\label{tab:arithscore}
\end{table}

\begin{table} [ht!]
\centering
\begin{tabular}{lcl}
\hline
\textbf{Prompt}& \textbf{Easy}&\textbf{Hard}\\
QAP25& 89.5& 24.3\\
QAP50& 87.7& 24.6\\ 
QAP100& 83.8& 26.9\\ 
QAP150& 81.4& 27.0\\
QAP200& 80.0& 25.0\\ 
TADB& 91.3& 20.3\\
 CoT& 85.8& 27.3\\
 PS+& 70.6& 21.1\\
\end{tabular}
\caption{Accuracy for Commonsense Reasoning}
\label{tab:score}
\end{table}

\subsection{Analysis of Word Count based on Question Difficulty }

Median word count generated by various prompts on all datasets and models categorized into easy and hard - where easy problems are those where baseline prompt leads to a correct answer and hard problems are those where baseline prompt leads to a wrong answer.

\begin{table}[ht!]
\centering
\begin{tabular}{lcl}
\hline
\textbf{Prompt}& \textbf{Easy}&\textbf{Hard}\\
QAP25& 94.6& 126.7\\
QAP50& 123.6& 158.5\\ 
QAP100& 200.4& 229.6\\ 
QAP150& 224.4& 257.9\\
QAP200& 270.0& 301.0\\ 
TADB& 146.3& 214.5\\
 CoT& 99.4& 128.3\\
 PS+& 197.8& 216.3\\
\end{tabular}
\caption{Mean word count for Arithmetic Reasoning}
\label{tab:arith}
\end{table}

\begin{table}[ht!]
\centering
\begin{tabular}{lcl}
\hline
\textbf{Prompt}& \textbf{Easy}&\textbf{Hard}\\
QAP25& 36.9& 38.7\\
QAP50& 71.5& 73.8\\ 
QAP100& 183.8& 192.3\\ 
QAP150& 215.8& 220.4\\
QAP200& 268.8& 274.6\\ 
TADB& 37.5& 58.0\\
 CoT& 29.1& 30.9\\
 PS+& 162.4& 179.0\\
\end{tabular}
\caption{Mean word count for Commonsense Reasoning}
\label{tab:count}
\end{table}

\pagebreak
\onecolumn

\subsection{Example Explanations}

\begin{figure*} [ht!]
    \centering
    \includegraphics[width=1\textwidth]{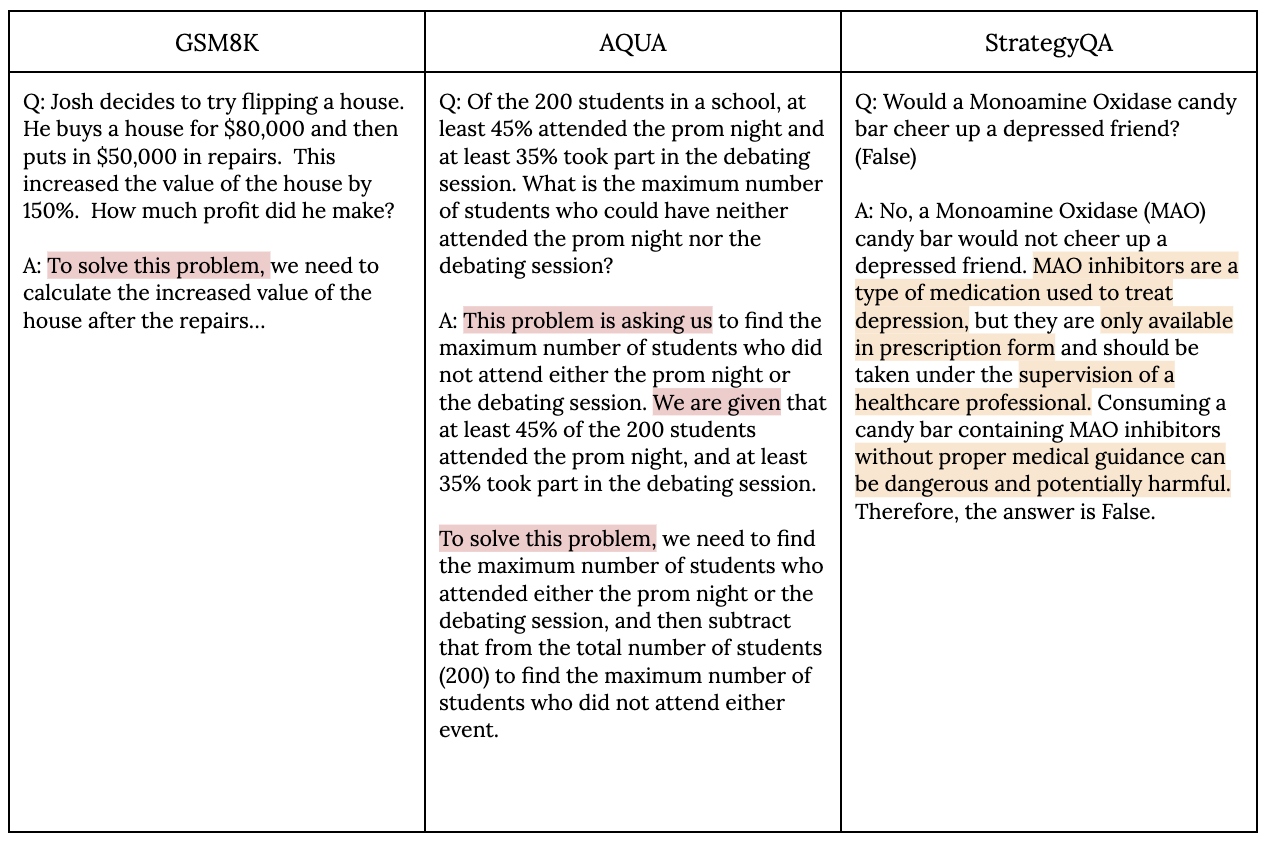}
    \captionsetup{width=1\textwidth}

    \caption{ Examples of QAP inducing explanations of the question on GSM8K, AQuA, and StrategyQA. The prompts include QAP50, QAP150, QAP50 respectively. Pink highlights key phrases (math reasoning) and orange highlights represents useful background information (commonsense reasoning).}
    \label{fig:enter-label}
\end{figure*}

\pagebreak
\subsection{Impact of Changing \textit{n}}
\begin{figure*} [htbp]
    \centering
    \includegraphics[width=1\linewidth]{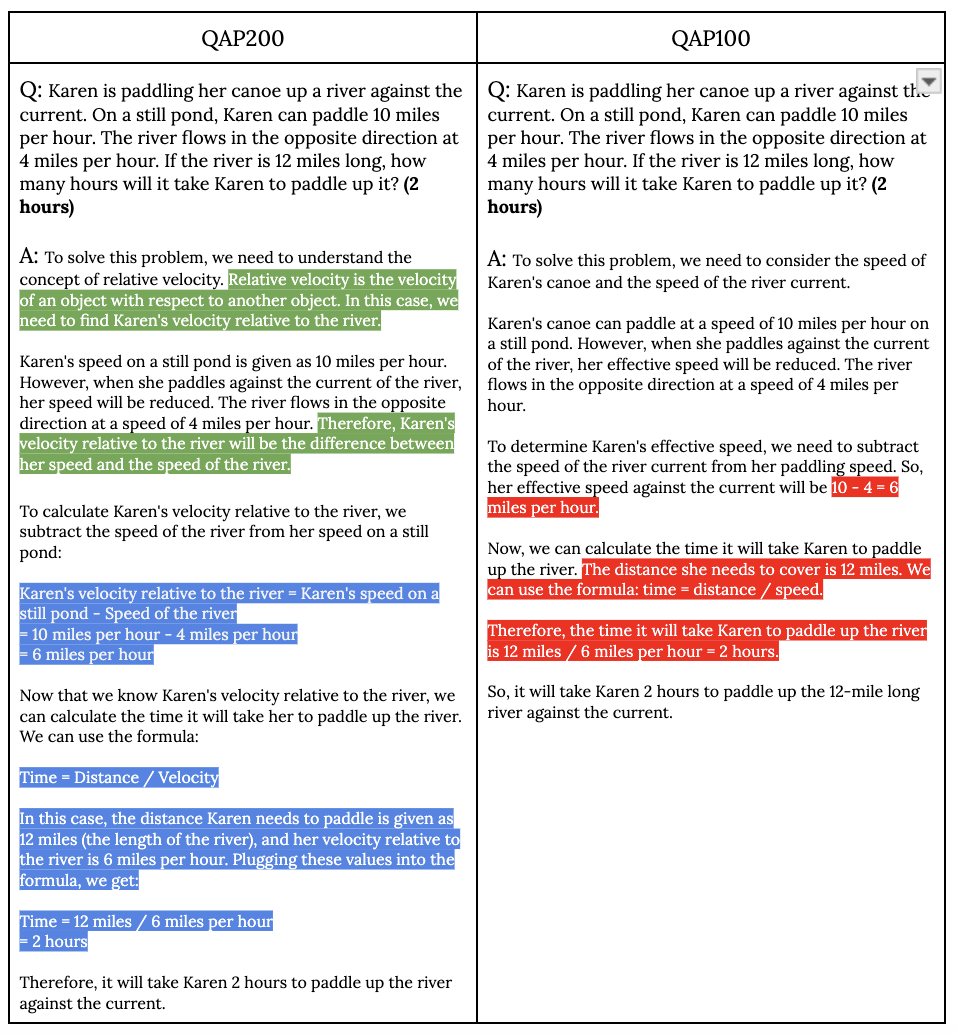}
    \caption{This comparison shows how responses vary when changing \textit{n}. This is only the answer portion. This was experimented on QAP50 and QAP20 on GSM8K on AQuA. \textbf{Blue} represents a QAP200 section which provides more detail than QAP100’s (\textbf{Red}) response on the same step. \textbf{Green} represents a section that QAP200 had that QAP100 did not have at all.}
    \label{fig:nexample}
\end{figure*}
\pagebreak

\subsection{Large value of n for simple problems hurts the performance}

\begin{figure*} [ht!]
    \centering
    \includegraphics[width=1\linewidth]{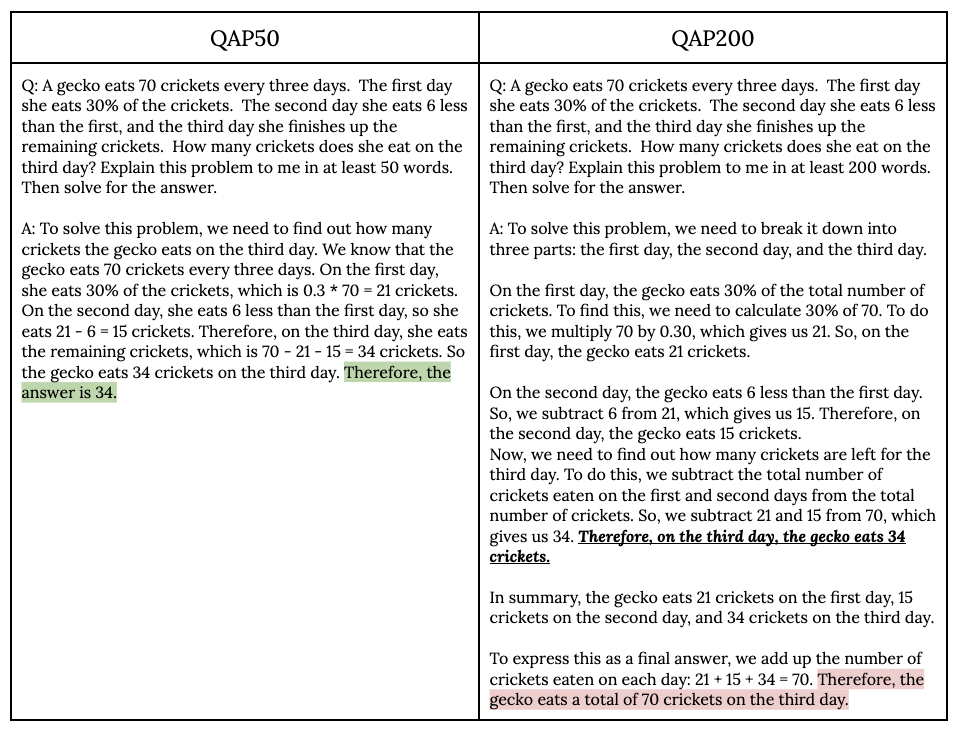}
    \caption{Example in which over-explanation can negatively impact a response. QAP50 acquires the correct answer (34), but QAP200 does not. In fact, QAP200 reaches the correct answer, but additional  explanation leads to a wrong answer.}
    \label{fig:OverEx}
\end{figure*}

\begin{figure*}
    \centering
    \includegraphics[width=1\linewidth]{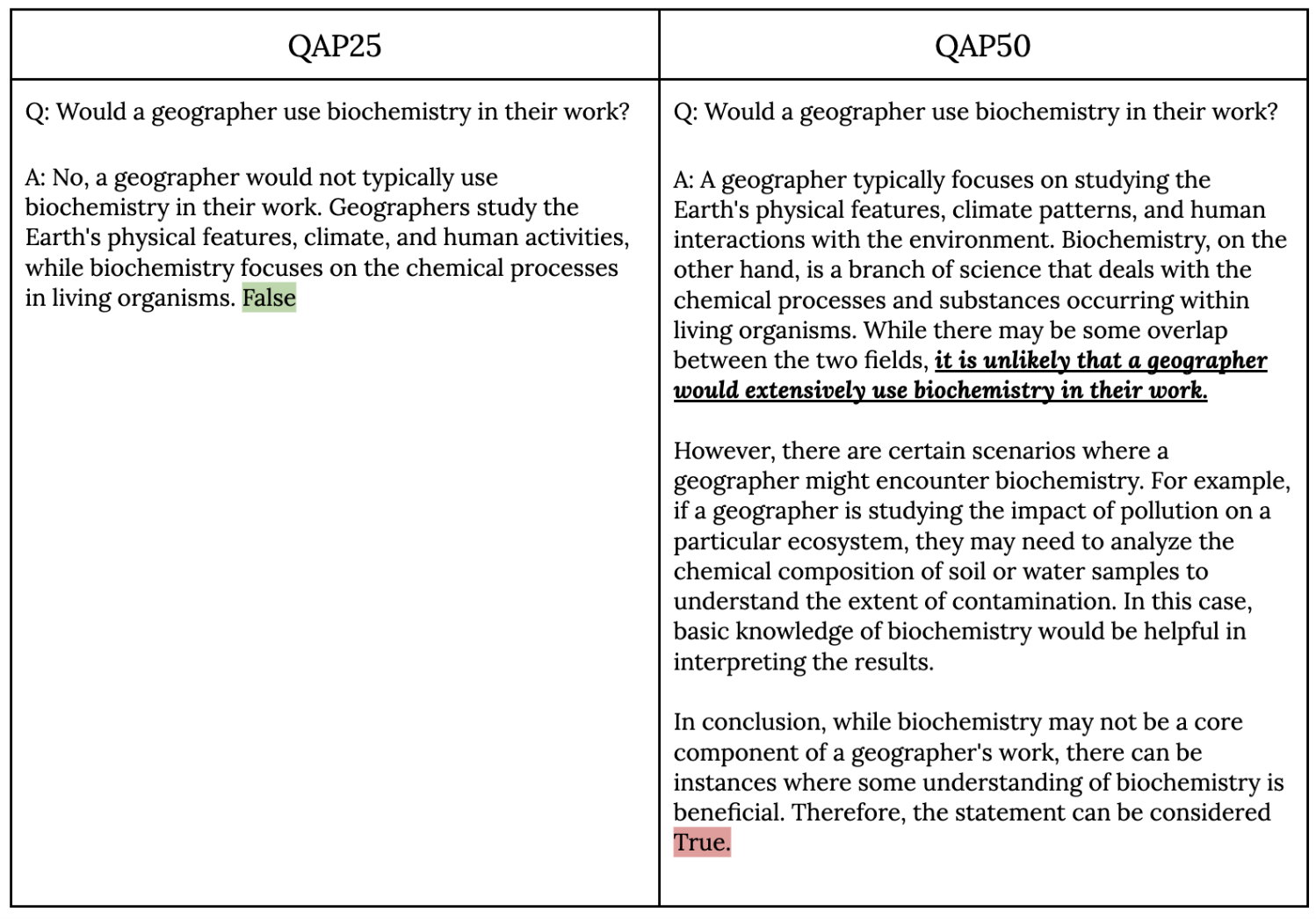}
    \caption{Example in which over-explanation negatively impacts a commonsense reasoning response. The comparison shows that more words can confuse the model. }
    \label{fig:comsenseoverex}
\end{figure*}

\pagebreak
\clearpage
\subsection{Word Counts for all datasets with  GPT-3.5 and GPT-4  }
\begin{figure*}[ht!]
    \centering
    \includegraphics[width=1\linewidth]{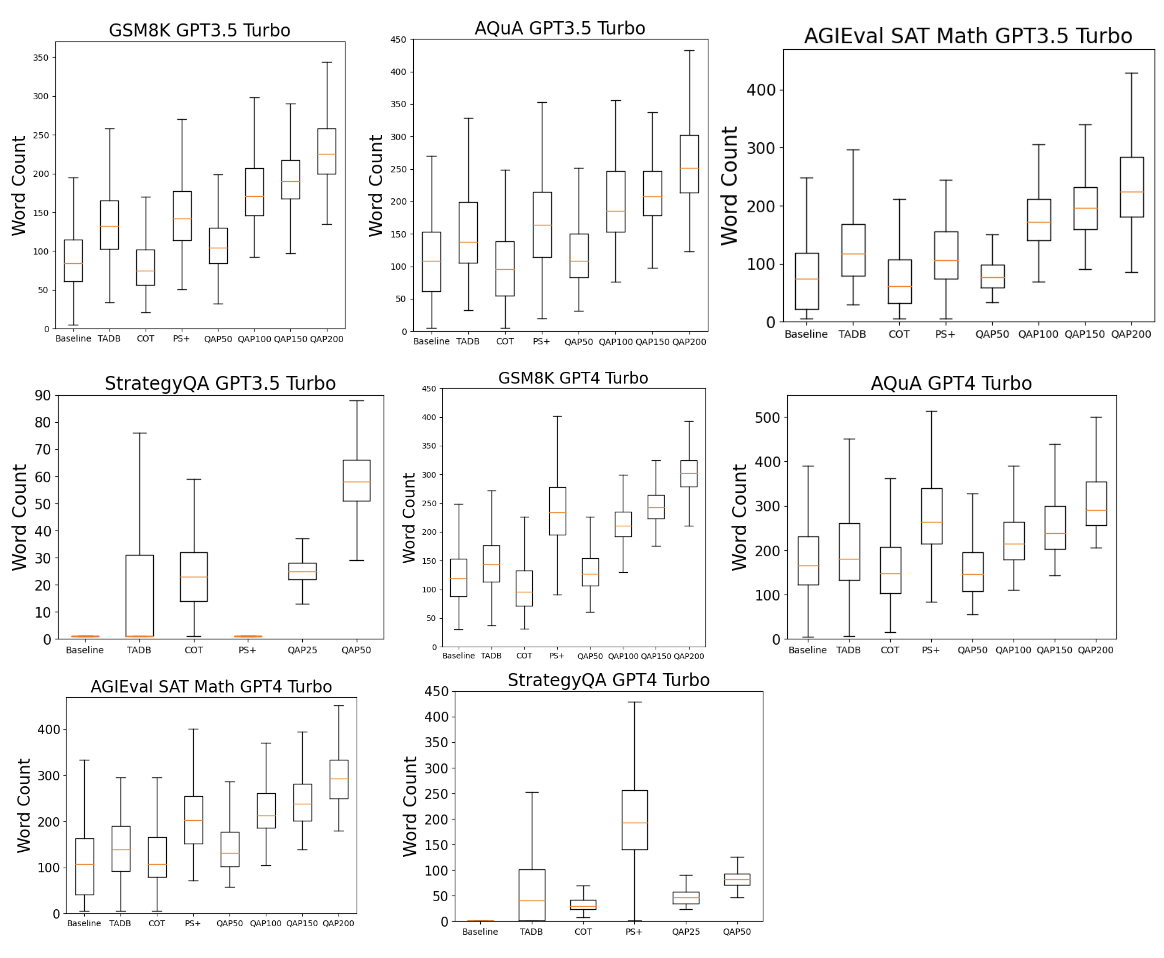}
    \caption{Median word counts in response for all datasets using GPT-3.5 Turbo and GPT-4 Turbo}
    \label{fig:boxplots}
\end{figure*}
\clearpage
\pagebreak
\subsection{QAP25 Unfinished Response}
\begin{figure*}[ht!]
    \centering
    \includegraphics[width=1\linewidth]{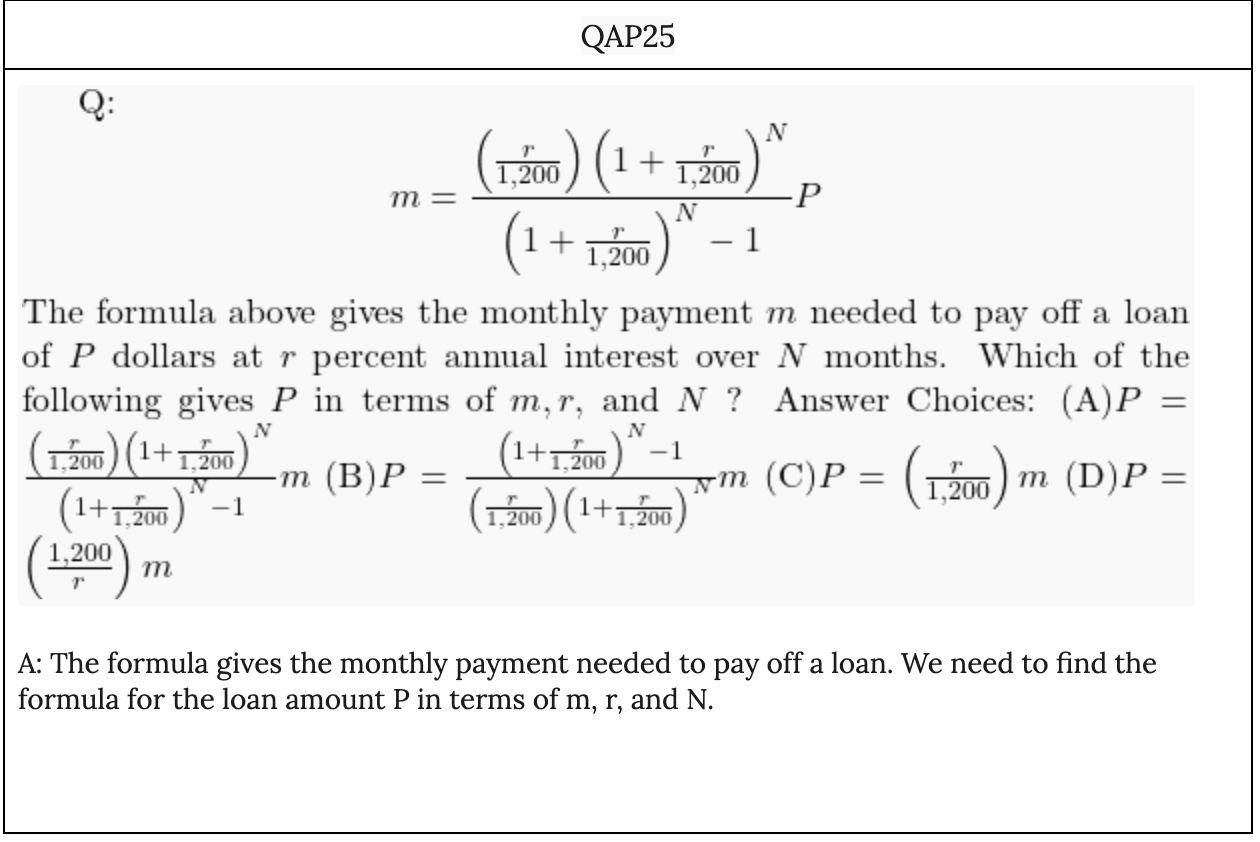}
    \caption{Example in which QAP25 outputs an unfinished response on the SAT dataset.}
    \label{fig:unfinresp}
\end{figure*}

\end{sloppypar}

\end{document}